\PassOptionsToPackage{unicode}{hyperref}
\PassOptionsToPackage{hyphens}{url}
\PassOptionsToPackage{dvipsnames,svgnames,x11names}{xcolor}
\documentclass[
  a4paper,
]{article}

\usepackage{amsmath,amssymb}
\usepackage{lmodern}
\usepackage{iftex}
\ifPDFTeX
  \usepackage[T1]{fontenc}
  \usepackage[utf8]{inputenc}
  \usepackage{textcomp} 
\else 
  \usepackage{unicode-math}
  \defaultfontfeatures{Scale=MatchLowercase}
  \defaultfontfeatures[\rmfamily]{Ligatures=TeX,Scale=1}
\fi
\IfFileExists{upquote.sty}{\usepackage{upquote}}{}
\IfFileExists{microtype.sty}{
  \usepackage[]{microtype}
  \UseMicrotypeSet[protrusion]{basicmath} 
}{}
\makeatletter
\@ifundefined{KOMAClassName}{
  \IfFileExists{parskip.sty}{%
    \usepackage{parskip}
  }{
    \setlength{\parindent}{0pt}
    \setlength{\parskip}{6pt plus 2pt minus 1pt}}
}{
  \KOMAoptions{parskip=half}}
\makeatother
\usepackage{xcolor}
\ifx\paragraph\undefined\else
  \let\oldparagraph\paragraph
  \renewcommand{\paragraph}[1]{\oldparagraph{#1}\mbox{}}
\fi
\ifx\subparagraph\undefined\else
  \let\oldsubparagraph\subparagraph
  \renewcommand{\subparagraph}[1]{\oldsubparagraph{#1}\mbox{}}
\fi

\usepackage{color}
\usepackage{fancyvrb}

\DefineVerbatimEnvironment{Highlighting}{Verbatim}{commandchars=\\\{\}}
\usepackage{framed}
\definecolor{shadecolor}{RGB}{46,52,64}
\newenvironment{Shaded}{\begin{snugshade}}{\end{snugshade}}

\newcommand{\BuiltInTok}[1]{\textcolor[rgb]{0.53,0.75,0.82}{\textit{#1}}}

\newcommand{\CommentTok}[1]{\textcolor[rgb]{0.38,0.43,0.53}{#1}}

\newcommand{\ControlFlowTok}[1]{\textcolor[rgb]{0.51,0.63,0.76}{\textbf{#1}}}

\newcommand{\DecValTok}[1]{\textcolor[rgb]{0.71,0.56,0.68}{#1}}

\newcommand{\FloatTok}[1]{\textcolor[rgb]{0.71,0.56,0.68}{#1}}
\newcommand{\FunctionTok}[1]{\textcolor[rgb]{0.53,0.75,0.82}{#1}}
\newcommand{\ImportTok}[1]{\textcolor[rgb]{0.64,0.75,0.55}{#1}}

\newcommand{\KeywordTok}[1]{\textcolor[rgb]{0.51,0.63,0.76}{\textbf{#1}}}
\newcommand{\NormalTok}[1]{\textcolor[rgb]{0.85,0.87,0.91}{#1}}
\newcommand{\OperatorTok}[1]{\textcolor[rgb]{0.51,0.63,0.76}{#1}}

\newcommand{\StringTok}[1]{\textcolor[rgb]{0.64,0.75,0.55}{#1}}
\newcommand{\VariableTok}[1]{\textcolor[rgb]{0.37,0.51,0.67}{#1}}

\providecommand{\tightlist}{%
  \setlength{\itemsep}{0pt}\setlength{\parskip}{0pt}}\usepackage{longtable,booktabs,array}
\usepackage{calc} 
\usepackage{etoolbox}
\makeatletter
\patchcmd\longtable{\par}{\if@noskipsec\mbox{}\fi\par}{}{}
\makeatother
\IfFileExists{footnotehyper.sty}{\usepackage{footnotehyper}}{\usepackage{footnote}}
\makesavenoteenv{longtable}
\usepackage{graphicx}
\makeatletter
\def\maxwidth{\ifdim\Gin@nat@width>\linewidth\linewidth\else\Gin@nat@width\fi}
\def\maxheight{\ifdim\Gin@nat@height>\textheight\textheight\else\Gin@nat@height\fi}
\makeatother
\setkeys{Gin}{width=\maxwidth,height=\maxheight,keepaspectratio}
\makeatletter
\def\fps@figure{htbp}
\makeatother
\newlength{\cslhangindent}
\newlength{\csllabelwidth}
\newlength{\cslentryspacingunit} 
\newenvironment{CSLReferences}[2] 
 {
  \setlength{\parindent}{0pt}
  \ifodd #1
  \let\oldpar\par
  \def\par{\hangindent=\cslhangindent\oldpar}
  \fi
  \setlength{\parskip}{#2\cslentryspacingunit}
 }%
 {}
\usepackage{calc}

\usepackage{arxiv}
\usepackage{orcidlink}
\usepackage{amsmath}
\usepackage[T1]{fontenc}
\makeatletter
\@ifpackageloaded{caption}{}{\usepackage{caption}}
\AtBeginDocument{%
\ifdefined\contentsname
  \renewcommand*\contentsname{Table of contents}
\else
  \newcommand\contentsname{Table of contents}
\fi
\ifdefined\listfigurename
  \renewcommand*\listfigurename{List of Figures}
\else
  \newcommand\listfigurename{List of Figures}
\fi
\ifdefined\listtablename
  \renewcommand*\listtablename{List of Tables}
\else
  \newcommand\listtablename{List of Tables}
\fi
\ifdefined\figurename
  \renewcommand*\figurename{Figure}
\else
  \newcommand\figurename{Figure}
\fi
\ifdefined\tablename
  \renewcommand*\tablename{Table}
\else
  \newcommand\tablename{Table}
\fi
}
\@ifpackageloaded{float}{}{\usepackage{float}}
\floatstyle{ruled}
\@ifundefined{c@chapter}{\newfloat{codelisting}{h}{lop}}{\newfloat{codelisting}{h}{lop}[chapter]}
\floatname{codelisting}{Listing}

\makeatother
\makeatletter
\@ifpackageloaded{caption}{}{\usepackage{caption}}
\@ifpackageloaded{subcaption}{}{\usepackage{subcaption}}
\makeatother
\ifLuaTeX
  \usepackage{selnolig}  
\fi
\IfFileExists{bookmark.sty}{\usepackage{bookmark}}{\usepackage{hyperref}}
\IfFileExists{xurl.sty}{\usepackage{xurl}}{} 
\hypersetup{
  pdftitle={Borch: A Deep Universal Probabilistic Programming Language},
  pdfauthor={Lewis Belcher; Johan Gudmundsson; Michael Green},
  pdfkeywords={uppl, neural networks, uncertainty, pytorch},
  colorlinks=true,
  linkcolor={blue},
  filecolor={Maroon},
  citecolor={Blue},
  urlcolor={Blue},
  pdfcreator={LaTeX via pandoc}}

\title{Borch: A Deep Universal Probabilistic Programming Language}
\author{
\textbf{Lewis Belcher}~\orcidlink{0000-0001-9680-078X}\\Department of AI
Research\\Desupervised
ApS\\Copenhagen,\ 2100\\\href{mailto:lb@desupervised.io}{lb@desupervised.io}\\\\\\
\textbf{Johan Gudmundsson}~\orcidlink{0000-0002-7316-0334}\\Department
of AI Research\\Desupervised
ApS\\Copenhagen,\ 2100\\\href{mailto:jg@desupervised.io}{jg@desupervised.io}\\\\\\
\textbf{Michael Green}~\orcidlink{0000-0003-1507-1613}\\Department of AI
Research\\Desupervised
ApS\\Copenhagen,\ 2100\\\href{mailto:mg@desupervised.io}{mg@desupervised.io}}
\date{}
\begin{document}
\maketitle
\begin{abstract}
Ever since the Multilayered Perceptron was first introduced the
connectionist community has struggled with the concept of uncertainty
and how this could be represented in these types of models. This past
decade has seen a lot of effort in trying to join the principled
approach of probabilistic modeling with the scalable nature of deep
neural networks. While the theoretical benefits of this consolidation
are clear, there are also several important practical aspects of these
endeavors; namely to force the models we create to represent, learn, and
report uncertainty in every prediction that is made. Many of these
efforts have been based on extending existing frameworks with additional
structures. We present Borch, a scalable deep universal probabilistic
programming language, built on top of PyTorch. The code is available for
download and use in our
\href{https://gitlab.com/desupervised/borch}{repo}.
\end{abstract}
{\bfseries \emph Keywords}
\def\sep{\textbullet\ }
uppl \sep neural networks \sep uncertainty \sep 
pytorch

\hypertarget{introduction}{%
\section{Introduction}\label{introduction}}

The ability to solve a wide variety of challenging real world problems
using machine learning has flourished during the course of the past
decade. We've seen advancements within diverse application areas, e.g.,
vision (Bojarski et al. 2016), natural language and physics (Bakarji et
al. 2022). We've also seen the emergence of a new paradigm for machine
learning where it is possible to teach a computer how to complete
mathematical proofs (Davis 2021; Davies et al. 2021) and even compete in
a real-world programming competition (Li et al. 2022). Despite the fact
that most of these advances were achieved by neural networks, there are
still areas where neural networks are far from being superior to more
traditional machine learning methods(Shwartz-Ziv and Armon 2021). The
strength in many of these methods lies in that they are easier to
interpret and reason about. However, despite these recent success, most
of the models developed and in use today lack the ability to adequately
quantify the confidence in the predictions given (Ghahramani 2015).

This is especially problematic for models that are exposed to out of
distribution (OOD) examples. When it comes to deploying neural networks
in real world applications, OOD data will inevitably be an issue
(Henriksson et al. 2021). Elaborate data augmentation processes and
larger training datasets can help make the networks more robust (Zheng
et al. 2016), but will never cover the entire spectrum of possible
scenarios. This is the most significant reason to model and report the
uncertainty in model parameters as well as the data.

With the ambition to understand uncertainty in predictions it is natural
to turn to a Bayesian treatment of model parameters, likelihood and data
(Dasgupta et al. 2020).

\hypertarget{sec-uncertainties}{%
\section{About Uncertainties}\label{sec-uncertainties}}

One of the most important concepts in modeling any system is to know
what you do not know (Kläs and Vollmer 2018). That may sound like an
impossible situation to resolve, but it's really quite simple. We don't
need to know why we don't know something, we only need to quantify the
degree of our ignorance. As such, it is important to categorize
different uncertainties to help us reason about where our lack of
knowledge resides.

\begin{figure}

{\centering 

\begin{figure}[H]

{\centering \includegraphics[width=6.5in,height=3.5in]{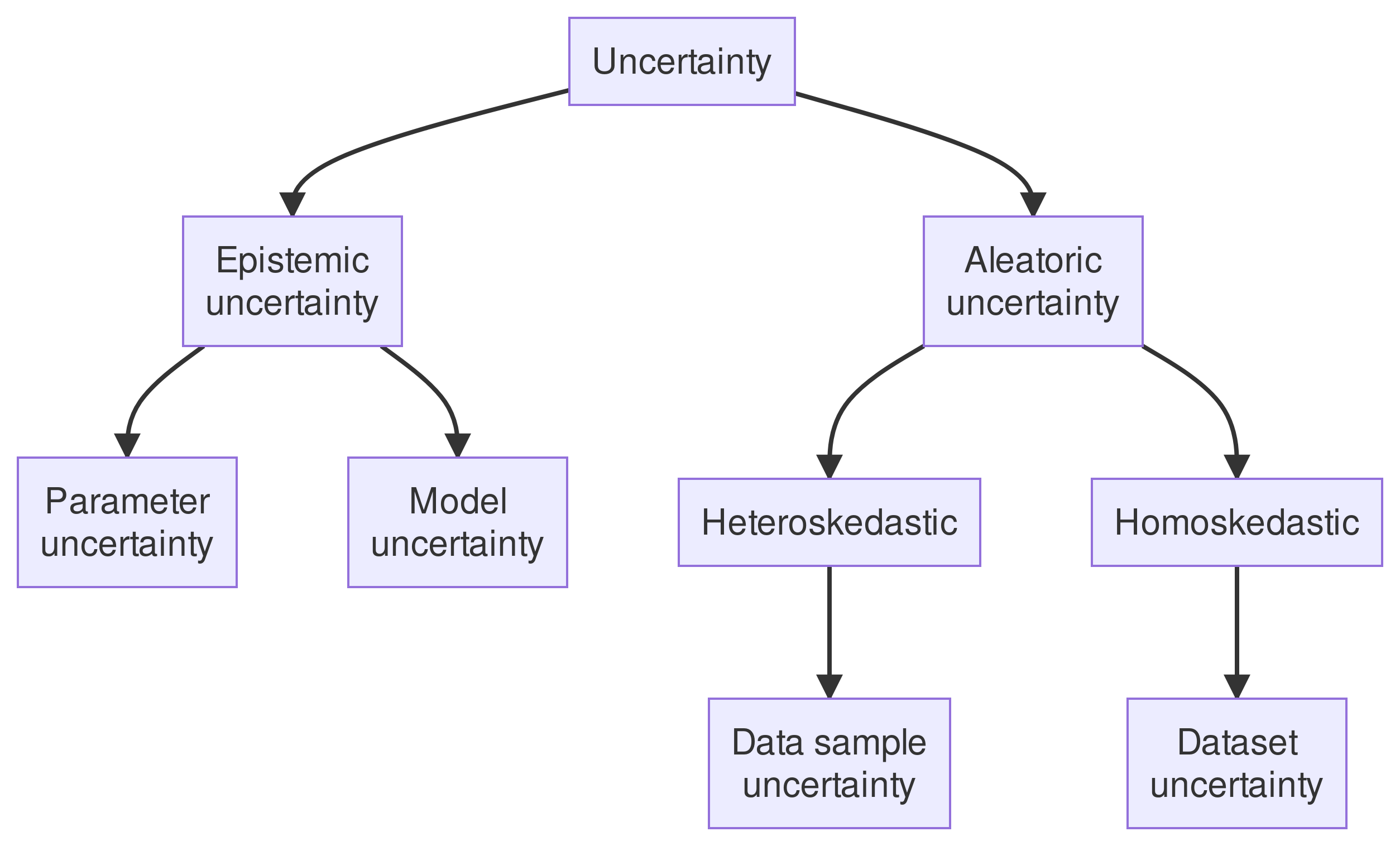}

}

\end{figure}

}

\caption{\label{fig-uncertainty}A classification of common uncertainties
that are useful to consider in machine learning models.}

\end{figure}

For most practical purposes in machine learning we categorize
uncertainties into Epistemic and Aleatoric uncertainty (Kiureghian and
Ditlevsen 2009). They are shown in Figure~\ref{fig-uncertainty}.

Epistemic uncertainty encapsulates the uncertainty the model has about
its own representation of the problem at hand. It represents all things
the model doesn't know but in principle could. Examples of this can be
data left out of the training set. Another example is structural
limitations of the model itself.

Aleatoric uncertainty on the other hand encapsulates the uncertainty
that happens as a consequence of running an experiment multiple times.
It's basically the innate uncertainty that we cannot model our way out
of. This particular type of uncertainty comes in two flavors.

Homoskedastic Aleatoric Uncertainty is the uncertainty or variance
observed across the entire dataset, i.e., one value.

Heteroskedastic Aleatoric Uncertainty on the other hand changes as a
function of each observation. This means that you can have a rather high
``data uncertainty'' for observation A, while having a low uncertainty
for observation B.

\hypertarget{probabilistic-programming}{%
\subsection{Probabilistic Programming}\label{probabilistic-programming}}

Universal Deep Probabilistic programming languages unify techniques for
formal description of computation with the representation and use of
uncertain knowledge (Goodman and Stuhlmüller 2014). These languages have
seen recent interest from artificial intelligence, programming
languages, cognitive science, and natural languages communities
(Gawlikowski et al. 2021).

Implementing probabilistic models from scratch is highly error-prone and
requires a high degree of expertise (Goodman and Stuhlmüller 2014;
Tolpin et al. 2016). This results in models that are very hard to
develop, extend and to reuse. To combat these issues practitioners
usually revert to using a probabilistic programming language (PPL) which
enables the expression of probabilistic programs and performing
inference on those programs.

We introduce Borch, a universal probabilistic programming language built
on PyTorch (Paszke et al. 2019).

Throughout the paper we illustrate code\footnote{The code snippets shown
  in the paper are just excerpts. Full examples are available at
  https://gitlab.com/desupervised/research/borch-paper and more
  elaborate tutorials are available at https://borch.readthedocs.io.} to
show the expressiveness of the language.

\hypertarget{design-principles}{%
\section{Design Principles}\label{design-principles}}

\hypertarget{basic-syntax}{%
\subsection{Basic Syntax}\label{basic-syntax}}

Borch is built around a few core primitives. (i) Distributions that
provide standard features such as generation of samples and calculation
of log probabilities of distributions etc. (ii) Random Variables (RV)
which unify distributions with tensors. (iii) Posteriors that handle how
the inference algorithm creates samples. (iv) A \texttt{Module} which is
the object used to keep track of RVs and the inference algorithm.

We recommend two different coding patters, functional and
object-oriented depending on the type of model used. The functional
syntax is straightforward and easy to understand. The forward pass of a
simple linear regression model can be implemented as follows:

\begin{Shaded}
\begin{Highlighting}[]
\KeywordTok{def}\NormalTok{ forward(bm, x):}
\NormalTok{    bm.b }\OperatorTok{=}\NormalTok{ Normal(}\DecValTok{0}\NormalTok{, }\DecValTok{3}\NormalTok{)}
\NormalTok{    bm.a }\OperatorTok{=}\NormalTok{ Normal(}\DecValTok{0}\NormalTok{, }\DecValTok{3}\NormalTok{)}
\NormalTok{    bm.sigma }\OperatorTok{=}\NormalTok{ HalfNormal(}\DecValTok{1}\NormalTok{)}
\NormalTok{    mu }\OperatorTok{=}\NormalTok{ bm.b }\OperatorTok{*}\NormalTok{ x }\OperatorTok{+}\NormalTok{ bm.a}
\NormalTok{    bm.y }\OperatorTok{=}\NormalTok{ Normal(mu, bm.sigma)}
    \ControlFlowTok{return}\NormalTok{ bm.y}
\end{Highlighting}
\end{Shaded}

The functional programming paradigm has the benefit of being more
similar to the actual mathematics of neural networks in flow and
execution order.

In some cases the object-oriented paradigm is more natural and the
corresponding code for the forward pass of the simple linear regression
model would look like below:

\begin{Shaded}
\begin{Highlighting}[]
\KeywordTok{class}\NormalTok{ Model(Module):}
    \KeywordTok{def} \FunctionTok{\_\_init\_\_}\NormalTok{(}\VariableTok{self}\NormalTok{):}
        \BuiltInTok{super}\NormalTok{().}\FunctionTok{\_\_init\_\_}\NormalTok{()}
        \VariableTok{self}\NormalTok{.b }\OperatorTok{=}\NormalTok{ Normal(}\DecValTok{0}\NormalTok{, }\DecValTok{3}\NormalTok{)}
        \VariableTok{self}\NormalTok{.a }\OperatorTok{=}\NormalTok{ Normal(}\DecValTok{0}\NormalTok{, }\DecValTok{3}\NormalTok{)}
        \VariableTok{self}\NormalTok{.sigma }\OperatorTok{=}\NormalTok{ HalfNormal(}\DecValTok{1}\NormalTok{)}

    \KeywordTok{def}\NormalTok{ forward(}\VariableTok{self}\NormalTok{, x):}
\NormalTok{        mu }\OperatorTok{=} \VariableTok{self}\NormalTok{.b }\OperatorTok{*}\NormalTok{ x }\OperatorTok{+} \VariableTok{self}\NormalTok{.a}
        \VariableTok{self}\NormalTok{.y }\OperatorTok{=}\NormalTok{ Normal(mu, }\VariableTok{self}\NormalTok{.sigma)}
        \ControlFlowTok{return} \VariableTok{self}\NormalTok{.y}
\end{Highlighting}
\end{Shaded}

For users of \texttt{torch.nn} (Paszke et al. 2019) and
\texttt{tensorflow.keras} (Chollet et al. 2015) this should look
familiar.

As illustrated in the code snippets above, Borch supports both
formalisms which enables flexibility in the model expression. Thus, one
can choose the most suitable approach for the model at hand. When
working with hierarchical models a functional style tends to produce
cleaner code (Buitléir, Russell, and Daly 2013) as there is a lot of
code that needs to be executed on each pass. When following the
object-oriented approach, the hierarchy should be expressed in the
forward pass, and only isolated random variables should be put in the
\texttt{\_\_init\_\_} function. This is a common pattern for neural
networks and thus the object-oriented paradigm is more compatible with
many existing tutorials and applications for neural networks.

In order to be an effective PPL it is important to have the ability to
condition on any random variable in the model. Further, being able to
fit a model while conditioning on an arbitrary set of random variables
is key. Also, un-conditioning a subset of those variables such the same
model can be used to generate predictions allows for the expression of a
wide variety of models. This is done in Borch using the \texttt{observe}
method on the \texttt{Module} object.

\begin{Shaded}
\begin{Highlighting}[]
\NormalTok{model.observe(y}\OperatorTok{=}\NormalTok{y)}
\NormalTok{forward(model, x) }\OperatorTok{==}\NormalTok{ y}
\end{Highlighting}
\end{Shaded}

Then in order to stop observing and be able to generate predictions from
the model one can use:

\begin{Shaded}
\begin{Highlighting}[]
\NormalTok{model.observe(}\VariableTok{None}\NormalTok{)}
\NormalTok{forward(model, x) }\OperatorTok{!=}\NormalTok{ y}
\end{Highlighting}
\end{Shaded}

This way of conditioning on data enables us to move from a generative
state to a discriminatory state in one line of code.

\hypertarget{scaling}{%
\subsection{Scaling}\label{scaling}}

In order to handle the ever-growing size of models and data, a PPL needs
to be scalable. This means handling everything from a simple linear
regression model to a billion parameter image classification neural
network trained on a graphics processing unit (GPU). At the same time it
needs to be flexible enough to express a wide range of models and that
these models can be fit with a multitude of inference methods. Borch is
composable such that models can be fit by the method most suitable to a
given problem. As such, any model can be conditioned on data through
diverse methods, e.g., Markov Chain Monte Carlo (MCMC) (Brooks et al.
2011), Variational Inference (VI) (Blei, Kucukelbir, and McAuliffe 2017)
and Maximum a Postiori (MAP) (Bassett and Deride 2018).

\hypertarget{variable-scoping}{%
\subsection{Variable Scoping}\label{variable-scoping}}

In many Python-based PPLs there is a lack of reflectiveness in the model
expression resulting in model specifications that require explicit
naming of the variables instantiated. For example in PyMC3 (Salvatier,
Wiecki, and Fonnesbeck 2016) it looks like this:

\begin{Shaded}
\begin{Highlighting}[]
\ControlFlowTok{with}\NormalTok{ pm.Model() }\ImportTok{as}\NormalTok{ linear\_model:}
\NormalTok{    weights }\OperatorTok{=}\NormalTok{ pm.Normal(}\StringTok{"weights"}\NormalTok{, mu}\OperatorTok{=}\DecValTok{0}\NormalTok{, sigma}\OperatorTok{=}\DecValTok{1}\NormalTok{)}
\end{Highlighting}
\end{Shaded}

and in Pyro (Bingham et al. 2019) it is written as below.

\begin{Shaded}
\begin{Highlighting}[]
\NormalTok{weights }\OperatorTok{=}\NormalTok{ pyro.sample(}\StringTok{"weights"}\NormalTok{, dist.Normal(}\DecValTok{0}\NormalTok{, }\DecValTok{1}\NormalTok{))}
\end{Highlighting}
\end{Shaded}

This results in the variable \texttt{weights} which will be shared
across the model and thus does not enable scoping using the standard
Python syntax but has to be handled manually or with custom primitives.
This lack of scoping results in a source of potential bugs, it is easy
to deal with for small models but as the model grows in complexity it
becomes more problematic. In Borch we strived to make probabilistic
programming as easy as we possibly could. That's why we can simply write
the following and avoid the verbose syntax present in, e.g., Pyro,
NumPyro and PyMC3.

\begin{Shaded}
\begin{Highlighting}[]
\NormalTok{module.weights }\OperatorTok{=}\NormalTok{ Normal(}\DecValTok{0}\NormalTok{, }\DecValTok{1}\NormalTok{)}
\end{Highlighting}
\end{Shaded}

\hypertarget{native-feel}{%
\subsection{Native Feel}\label{native-feel}}

We aim to allow the full expressiveness of normal Python code and ease
of integration with other tools. Since PyTorch (Paszke et al. 2019)
offers auto differentiation for dynamic computational graphs, GPU
support and native integration with Python primitives the same holds
true for Borch since it builds on top of PyTorch. Thus, we support
features like stochastic control flow out of the box.

\begin{Shaded}
\begin{Highlighting}[]
\KeywordTok{def}\NormalTok{ forward(module, data):}
    \ControlFlowTok{if}\NormalTok{ randn(}\DecValTok{1}\NormalTok{) }\OperatorTok{\textgreater{}} \DecValTok{0}\NormalTok{:}
\NormalTok{        module.weight }\OperatorTok{=}\NormalTok{ Normal(}\OperatorTok{{-}}\DecValTok{1}\NormalTok{, }\DecValTok{1}\NormalTok{)}
    \ControlFlowTok{else}\NormalTok{:}
\NormalTok{        module.weight }\OperatorTok{=}\NormalTok{ Normal(}\DecValTok{1}\NormalTok{, }\DecValTok{10}\NormalTok{)}
    \ControlFlowTok{return}\NormalTok{ data}\OperatorTok{*}\NormalTok{torch.exp(module.weight)}
\end{Highlighting}
\end{Shaded}

\hypertarget{neural-networks}{%
\subsection{Neural Networks}\label{neural-networks}}

We set out to keep the feel of how PyTorch expresses neural networks and
make it a seamless experience to do Bayesian inference using Borch for
anyone with PyTorch experience. Thus, we wanted it to be fully
compatible with normal Python functions. We also wanted a tight
integration with PyTorch such that one gets dynamic control flow and the
same type of user experience when creating models in Borch.

To enable this for neural networks we provide an interface almost
identical to the \texttt{torch.nn} modules and in most cases it is
possible to just switch from \texttt{import\ torch.nn\ as\ nn} to
\texttt{import\ borch.nn\ as\ nn}. This makes a network defined in
PyTorch probabilistic, without any other changes in the model
specification. Thus, a simple MNIST (Deng 2012) image classification
model can be written like below.

\begin{Shaded}
\begin{Highlighting}[]
\ImportTok{import}\NormalTok{ torch.nn.functional }\ImportTok{as}\NormalTok{ F}
\ImportTok{from}\NormalTok{ borch }\ImportTok{import}\NormalTok{ nn}

\KeywordTok{class}\NormalTok{ Net(nn.Module):}
    \KeywordTok{def} \FunctionTok{\_\_init\_\_}\NormalTok{(}\VariableTok{self}\NormalTok{):}
        \BuiltInTok{super}\NormalTok{(Net, }\VariableTok{self}\NormalTok{).}\FunctionTok{\_\_init\_\_}\NormalTok{()}
        \VariableTok{self}\NormalTok{.conv1 }\OperatorTok{=}\NormalTok{ nn.Conv2d(}\DecValTok{1}\NormalTok{, }\DecValTok{6}\NormalTok{, }\DecValTok{5}\NormalTok{)}
        \VariableTok{self}\NormalTok{.conv2 }\OperatorTok{=}\NormalTok{ nn.Conv2d(}\DecValTok{6}\NormalTok{, }\DecValTok{16}\NormalTok{, }\DecValTok{5}\NormalTok{)}
        \VariableTok{self}\NormalTok{.fc1 }\OperatorTok{=}\NormalTok{ nn.Linear(}\DecValTok{16} \OperatorTok{*} \DecValTok{5} \OperatorTok{*} \DecValTok{5}\NormalTok{, }\DecValTok{120}\NormalTok{)}
        \VariableTok{self}\NormalTok{.fc2 }\OperatorTok{=}\NormalTok{ nn.Linear(}\DecValTok{120}\NormalTok{, }\DecValTok{84}\NormalTok{)}
        \VariableTok{self}\NormalTok{.fc3 }\OperatorTok{=}\NormalTok{ nn.Linear(}\DecValTok{84}\NormalTok{, }\DecValTok{10}\NormalTok{)}

    \KeywordTok{def}\NormalTok{ forward(}\VariableTok{self}\NormalTok{, x):}
\NormalTok{        x }\OperatorTok{=}\NormalTok{ F.max\_pool2d(F.relu(}\VariableTok{self}\NormalTok{.conv1(x)), (}\DecValTok{2}\NormalTok{, }\DecValTok{2}\NormalTok{))}
\NormalTok{        x }\OperatorTok{=}\NormalTok{ F.max\_pool2d(F.relu(}\VariableTok{self}\NormalTok{.conv2(x)), }\DecValTok{2}\NormalTok{)}
\NormalTok{        x }\OperatorTok{=}\NormalTok{ x.view(}\DecValTok{1}\NormalTok{, }\OperatorTok{{-}}\DecValTok{1}\NormalTok{)}
\NormalTok{        x }\OperatorTok{=}\NormalTok{ F.relu(}\VariableTok{self}\NormalTok{.fc1(x))}
\NormalTok{        x }\OperatorTok{=}\NormalTok{ F.relu(}\VariableTok{self}\NormalTok{.fc2(x))}
\NormalTok{        x }\OperatorTok{=} \VariableTok{self}\NormalTok{.fc3(x)}
        \ControlFlowTok{return}\NormalTok{ x}
\end{Highlighting}
\end{Shaded}

Changing an entire neural network to be Bayesian is not always required
though. Often it makes sense to only make specific parts or layers of
the neural network Bayesian. Using the MNIST example above we can easily
compose Borch and PyTorch such that only the two last layers are
Bayesian. This means that the last decision layers will learn
uncertainty while the earlier layers of the network will have point
parameters. We write the adapted MNIST model as below.

\begin{Shaded}
\begin{Highlighting}[]
\ImportTok{import}\NormalTok{ torch}
\ImportTok{import}\NormalTok{ borch}
\ImportTok{import}\NormalTok{ torch.nn.functional }\ImportTok{as}\NormalTok{ F}

\KeywordTok{class}\NormalTok{ Net(borch.nn.Module):}
    \KeywordTok{def} \FunctionTok{\_\_init\_\_}\NormalTok{(}\VariableTok{self}\NormalTok{):}
        \BuiltInTok{super}\NormalTok{ (Net, }\VariableTok{self}\NormalTok{).}\FunctionTok{\_\_init\_\_}\NormalTok{()}
        \VariableTok{self}\NormalTok{.conv1 }\OperatorTok{=}\NormalTok{ torch.nn.Conv2d(}\DecValTok{1}\NormalTok{, }\DecValTok{6}\NormalTok{, }\DecValTok{5}\NormalTok{)}
        \VariableTok{self}\NormalTok{.conv2 }\OperatorTok{=}\NormalTok{ torch.nn.Conv2d(}\DecValTok{6}\NormalTok{, }\DecValTok{16}\NormalTok{, }\DecValTok{5}\NormalTok{)}
        \VariableTok{self}\NormalTok{.fc1 }\OperatorTok{=}\NormalTok{ torch.nn.Linear(}\DecValTok{16} \OperatorTok{*} \DecValTok{5} \OperatorTok{*} \DecValTok{5}\NormalTok{, }\DecValTok{120}\NormalTok{)}
        \VariableTok{self}\NormalTok{.fc2 }\OperatorTok{=}\NormalTok{ borch.nn.Linear(}\DecValTok{120}\NormalTok{, }\DecValTok{84}\NormalTok{)}
        \VariableTok{self}\NormalTok{.fc3 }\OperatorTok{=}\NormalTok{ borch.nn.Linear(}\DecValTok{84}\NormalTok{, }\DecValTok{10}\NormalTok{)}
    
    \KeywordTok{def}\NormalTok{ forward(}\VariableTok{self}\NormalTok{, x):}
\NormalTok{        x }\OperatorTok{=}\NormalTok{ F.max\_pool2d(F.relu(}\VariableTok{self}\NormalTok{.conv1(x)), (}\DecValTok{2}\NormalTok{, }\DecValTok{2}\NormalTok{))}
\NormalTok{        x }\OperatorTok{=}\NormalTok{ F.max\_pool2d(F.relu(}\VariableTok{self}\NormalTok{.conv2(x)), }\DecValTok{2}\NormalTok{)}
\NormalTok{        x }\OperatorTok{=}\NormalTok{ x.view(}\DecValTok{1}\NormalTok{, }\OperatorTok{{-}}\DecValTok{1}\NormalTok{)}
\NormalTok{        x }\OperatorTok{=}\NormalTok{ F.relu(}\VariableTok{self}\NormalTok{.fc1(x))}
\NormalTok{        x }\OperatorTok{=}\NormalTok{ F.relu(}\VariableTok{self}\NormalTok{.fc2(x))}
\NormalTok{        x }\OperatorTok{=} \VariableTok{self}\NormalTok{.fc3(x)}
        \ControlFlowTok{return}\NormalTok{ x}
\end{Highlighting}
\end{Shaded}

\begin{figure}

{\centering \includegraphics{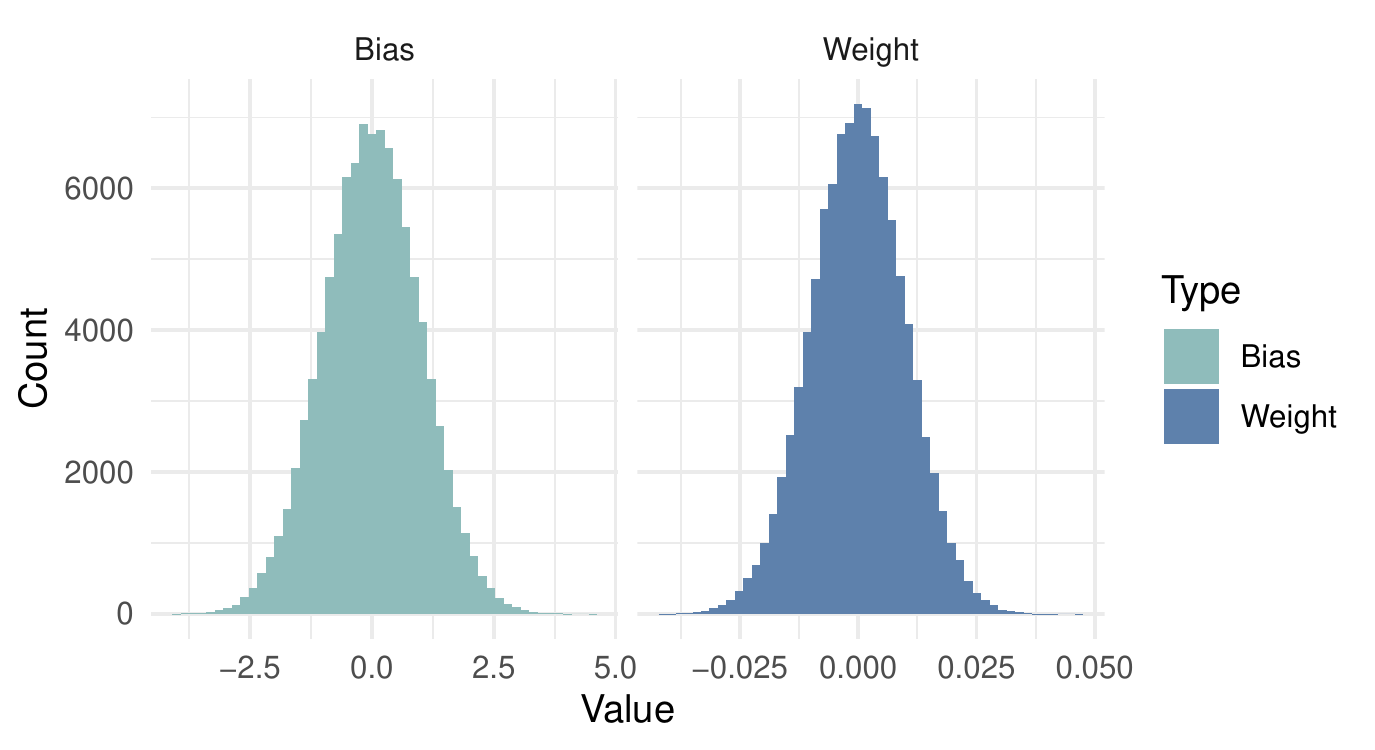}

}

\caption{\label{fig-mnistpriors}Illustration of the histograms of the
two prior distributions broadcast to the weights and the biases of the
dense layer \texttt{fc3} in the MNIST model. In this example the biases
and weights are given by \(b \sim N(0, 1)\) and
\(\omega \sim N(0, 0.01)\) respectively.}

\end{figure}

The signature for PyTorch \texttt{nn} modules has also been extended to
support the ability to set priors for the weights of an \texttt{nn}
module. Thus, to use the priors in Figure~\ref{fig-mnistpriors} for the
fully connected dense layer \texttt{fc3} in the Bayesian MNIST model, it
is as simple as

\begin{Shaded}
\begin{Highlighting}[]
\VariableTok{self}\NormalTok{.fc3 }\OperatorTok{=}\NormalTok{ nn.Linear(}\DecValTok{84}\NormalTok{, }\DecValTok{10}\NormalTok{, weight}\OperatorTok{=}\NormalTok{Normal(}\DecValTok{0}\NormalTok{, }\FloatTok{1e{-}2}\NormalTok{), bias}\OperatorTok{=}\NormalTok{Normal(}\DecValTok{0}\NormalTok{, }\DecValTok{1}\NormalTok{))}
\end{Highlighting}
\end{Shaded}

where the weight and bias arguments are broadcast to the shape of each
tensor in the \texttt{nn.Linear} module.

\hypertarget{transforming-networks-to-become-bayesian}{%
\subsection{Transforming Networks to Become
Bayesian}\label{transforming-networks-to-become-bayesian}}

Transfer learning (Farahani et al. 2021) is commonly used to take a
network trained on one problem as a starting point and train it to solve
a different problem. In the same way Borch allows taking a
\texttt{torch.nn} network and converting the whole or a part of it to be
Bayesian. This makes it possible to quickly introduce uncertainty to a
full scale model.

Here we show how to create a Bayesian MobileNetV3 (Howard et al. 2019)
network in just a few lines of code.

\begin{Shaded}
\begin{Highlighting}[]
\ImportTok{import}\NormalTok{ torchvision}
\ImportTok{import}\NormalTok{ borch}
\ImportTok{import}\NormalTok{ torch}
\NormalTok{mobilenet }\OperatorTok{=}\NormalTok{ borch.nn.borchify\_network(}
\NormalTok{    torchvision.models.mobilenet\_v3\_small(pretrained}\OperatorTok{=}\VariableTok{True}\NormalTok{),}
\NormalTok{)}
\CommentTok{\# Now it can be used to predict}
\NormalTok{borch.sample(mobilenet)}
\NormalTok{mobilenet(torch.randn(}\DecValTok{3}\NormalTok{, }\DecValTok{3}\NormalTok{, }\DecValTok{224}\NormalTok{, }\DecValTok{224}\NormalTok{))}
\end{Highlighting}
\end{Shaded}

However, we do not need to make the entire network Bayesian. Instead, we
can make parts of the network Bayesian and the process is the same.

\begin{Shaded}
\begin{Highlighting}[]
\NormalTok{mobilenet}\OperatorTok{=}\NormalTok{ models.mobilenet\_v3\_small(pretrained}\OperatorTok{=}\VariableTok{True}\NormalTok{)}
\NormalTok{mobilenet.classifier }\OperatorTok{=}\NormalTok{ borchify\_network(mobilenet.classifier)}
\NormalTok{borch.sample(mobilenet)}
\NormalTok{mobilenet(torch.randn(}\DecValTok{3}\NormalTok{, }\DecValTok{3}\NormalTok{, }\DecValTok{224}\NormalTok{, }\DecValTok{224}\NormalTok{))}
\end{Highlighting}
\end{Shaded}

A couple of notable differences compared to PyTorch are:

\begin{enumerate}
\def\labelenumi{\arabic{enumi}.}
\tightlist
\item
  We need to \texttt{borchify} or \texttt{lift} the network up to a
  probabilistic network.
\item
  We need to sample the posterior weights before generating a
  prediction. Without the sample statement we will reuse the same
  weights over and over again.
\end{enumerate}

Note that it makes no difference whether we call \texttt{borch.sample}
on only the classifier (\texttt{borch.sample(mobilenet.classifier)}) or
on the entire network (\texttt{borch.sample(mobilenet)}) as the
operation is applied recursively and only affects the segments of the
network which are constructed with Borch.

\hypertarget{inference}{%
\subsection{Inference}\label{inference}}

The primary focus in the development of Borch has been to ensure that
one can actually create and train state of the art neural networks. In
order to achieve this a lot of attention has been focused on variational
inference (Paisley, Blei, and Jordan 2012) as an inference method, as
it, together with MAP, are the fitting methods that can scale to big
neural networks.

To this end we wanted to make it easy to control the approximating
distribution for different parts of the network. This is done by the
concept of a posterior, that guides the approximating distribution. In
order to offer full control of the approximating distribution we have
\texttt{Manual} posterior. Below is a minimal example where we simply
infer the location and scale of a normal distribution.

\begin{Shaded}
\begin{Highlighting}[]
\KeywordTok{def}\NormalTok{ forward(mod):}
\NormalTok{    mod.test }\OperatorTok{=}\NormalTok{ Normal(}\DecValTok{5}\NormalTok{, }\DecValTok{1}\NormalTok{)}
\end{Highlighting}
\end{Shaded}

The posterior is specified in the same way. In order for the parameters
to be learnable we need to define them as \texttt{torch.Parameters}. The
\texttt{Parameter} wrapper for tensors just lets the framework know that
it should be returned when calling \texttt{model.parameters()} in the
same way \texttt{torch} does it. Thus, this is a convenient way to pass
them to an optimizer.

\begin{Shaded}
\begin{Highlighting}[]
\NormalTok{man\_posterior }\OperatorTok{=}\NormalTok{ posterior.Manual()}
\NormalTok{man\_posterior.mean }\OperatorTok{=}\NormalTok{ torch.nn.Parameter(torch.ones(}\DecValTok{1}\NormalTok{))}
\NormalTok{man\_posterior.sd }\OperatorTok{=}\NormalTok{ torch.nn.Parameter(torch.ones(}\DecValTok{1}\NormalTok{))}

\KeywordTok{def}\NormalTok{ forward\_posterior(posterior):}
\NormalTok{    scale }\OperatorTok{=}\NormalTok{ torch.exp(posterior.sd)}\OperatorTok{+}\FloatTok{0.01}
\NormalTok{    mean }\OperatorTok{=}\NormalTok{ posterior.mean.}\BuiltInTok{abs}\NormalTok{()}
\NormalTok{    posterior.test }\OperatorTok{=}\NormalTok{ dist.Normal(mean, scale)}
\end{Highlighting}
\end{Shaded}

While we support the manual expression of the posterior using the
\texttt{borch.posterior.Manual}, we found that in the case of large
neural networks it was simply easier to automatically create the
appropriate approximate posterior. This still allows control over what
posterior is used where in the network. It is done by simply specifying
the posterior for each module.

\begin{Shaded}
\begin{Highlighting}[]
\KeywordTok{class}\NormalTok{ Net(nn.Module):}
    \KeywordTok{def} \FunctionTok{\_\_init\_\_}\NormalTok{(}\VariableTok{self}\NormalTok{):}
        \BuiltInTok{super}\NormalTok{(Net, }\VariableTok{self}\NormalTok{).}\FunctionTok{\_\_init\_\_}\NormalTok{(posterior}\OperatorTok{=}\NormalTok{borch.posterior.Automatic())}
        \VariableTok{self}\NormalTok{.conv1 }\OperatorTok{=}\NormalTok{ Conv2d(}\DecValTok{1}\NormalTok{, }\DecValTok{6}\NormalTok{, }\DecValTok{5}\NormalTok{, posterior}\OperatorTok{=}\NormalTok{borch.posterior.Normal(log\_scale}\OperatorTok{={-}}\DecValTok{3}\NormalTok{))}
        \VariableTok{self}\NormalTok{.conv2 }\OperatorTok{=}\NormalTok{ Conv2d(}\DecValTok{6}\NormalTok{, }\DecValTok{16}\NormalTok{, }\DecValTok{5}\NormalTok{, posterior}\OperatorTok{=}\NormalTok{borch.posterior.Normal(log\_scale}\OperatorTok{={-}}\DecValTok{3}\NormalTok{))}
        \VariableTok{self}\NormalTok{.fc1 }\OperatorTok{=}\NormalTok{ Linear(}\DecValTok{16} \OperatorTok{*} \DecValTok{5} \OperatorTok{*} \DecValTok{5}\NormalTok{, }\DecValTok{120}\NormalTok{,}
\NormalTok{                          posterior}\OperatorTok{=}\NormalTok{borch.posterior.ScaledNormal(scaling}\OperatorTok{=}\FloatTok{1e{-}2}\NormalTok{)}
\NormalTok{                         )}
        \VariableTok{self}\NormalTok{.fc2 }\OperatorTok{=}\NormalTok{ Linear(}\DecValTok{120}\NormalTok{, }\DecValTok{10}\NormalTok{,}
\NormalTok{                          posterior}\OperatorTok{=}\NormalTok{borch.posterior.ScaledNormal(scaling}\OperatorTok{=}\FloatTok{1e{-}2}\NormalTok{)}
\NormalTok{                         )}
    \KeywordTok{def}\NormalTok{ forward(}\VariableTok{self}\NormalTok{, x):}
\NormalTok{        x }\OperatorTok{=}\NormalTok{ F.max\_pool2d(F.relu(}\VariableTok{self}\NormalTok{.conv1(x)), (}\DecValTok{2}\NormalTok{, }\DecValTok{2}\NormalTok{))}
\NormalTok{        x }\OperatorTok{=}\NormalTok{ F.max\_pool2d(F.relu(}\VariableTok{self}\NormalTok{.conv2(x)), }\DecValTok{2}\NormalTok{)}
\NormalTok{        x }\OperatorTok{=}\NormalTok{ x.view(}\DecValTok{1}\NormalTok{, }\OperatorTok{{-}}\DecValTok{1}\NormalTok{)}
\NormalTok{        x }\OperatorTok{=}\NormalTok{ F.relu(}\VariableTok{self}\NormalTok{.fc1(x))}
\NormalTok{        x }\OperatorTok{=} \VariableTok{self}\NormalTok{.fc2(x)}
        \VariableTok{self}\NormalTok{.classification }\OperatorTok{=}\NormalTok{ Categorical(logits}\OperatorTok{=}\NormalTok{x)}
        \ControlFlowTok{return}\NormalTok{ x}
\end{Highlighting}
\end{Shaded}

In this example we used the \texttt{Automatic} posterior for the top
level module as it will automatically detect that the \texttt{x} in
\texttt{self.classification\ =\ Categorical(logits=x)} will be updated
in each forward. Thus, it will not create a parameter for
\texttt{logits} that would be optimized. For the layers \texttt{conv1}
and \texttt{conv2} the \texttt{Normal} posterior was used, creating a
normal distribution as an approximating distribution where the scale
goes through a \texttt{exp} transform. We can control at what value the
unconstrained scale is initialized at. This is different to the
\texttt{ScaledNormal} used for \texttt{fc1} and \texttt{fc2} which
creates a Normal distribution as the approximating distribution but
scales the width of the initialization point based on the prior.

When using a PPL, depending on the model, one will be interested in
using different inference methods. For high confidence in the accuracy
of the fit posteriors MCMC methods like The No-U-Turn Sampler (Hoffman
and Gelman 2011) are preferred. For bigger models that are more data
hungry like a neural network, MCMC methods do not scale well and also
suffer from sampling issues (Blei, Kucukelbir, and McAuliffe 2017;
Jospin et al. 2022) due to a massively degenerate energy landscape.
Here, we need to use some other method like MAP or VI (Blei, Kucukelbir,
and McAuliffe 2017).

Borch enables this by tying inference methods to a specific posterior.
Thus, if we want to fit a model with MCMC even though the model set a
posterior tied to VI at creation, we can simply replace the posterior on
all the modules in a model like this:

\begin{Shaded}
\begin{Highlighting}[]
\NormalTok{net }\OperatorTok{=}\NormalTok{ Net()}
\NormalTok{net.}\BuiltInTok{apply}\NormalTok{(set\_posteriors(PointMass))}
\end{Highlighting}
\end{Shaded}

This facilitates the ability to encode the desired posterior when
expressing the model. It also allows us to quickly alter the posterior
on an instantiated object in order to use a different inference method,
or simply to test the effect of a different posterior guide. The use of
the term \texttt{PointMass} can be confusing but here it refers to a
single sample, since each forward pass produces a set of values. It is
the distribution of many forward passes that will eventually make up the
posterior in this case.

\hypertarget{comparison-to-existing-ppls}{%
\section{Comparison to Existing
PPLs}\label{comparison-to-existing-ppls}}

The development of Borch started in 2018 out of a necessity to build
robust deep learning models that perform well on OOD data. The models
developed at Desupervised were required to perform in critical
environments where the consequence of a bad decision could be
devastating. With this particular need, we realized that a Bayesian
formalism made the most sense, but we were unhappy with the performance
and coding style of existing frameworks. So we wanted to build large and
deep neural networks while still maintaining a standard PPL interface.
Thus, Borch is by no means the first PPL of its kind since there has
been active development in this area for a long time. The landscape
today is also different from when Borch was first developed.

Today, in our point of view, the two main alternatives are Pyro (Bingham
et al. 2019) and Tensorflow probability (Dillon et al. 2017). Pyro
offers a very solid PPL but very little tooling for neural networks.
Tensorflow probability integrates with the Tensorflow Keras api that
allows the creation of neural networks in a similar fashion as one can
do with Borch.

Borch offers a wider range of Neural Network modules and an interface as
a universal PPL that is more intuitive and does not require manual
creation of the joint distribution of the model. Several PPLs that
influenced the design of Borch were Stan (Stan Development Team 2018),
Edward (Tran et al. 2016), PyMC3 (Salvatier, Wiecki, and Fonnesbeck
2016) and ProbTorch (Siddharth et al. 2017).

\hypertarget{conclusion}{%
\section{Conclusion}\label{conclusion}}

In order for deep neural networks to be deployed and used at scale
across safety critical industries, e.g., healthcare, pharma, and
construction they need the ability to report uncertainty for every
prediction made. The importance does not reside in the pursuit of
statistical excellence but rather in the necessity for the end user to
know how much confidence should be put into the prediction at hand. The
likelihood is of little help here since it's uncalibrated, unnormalized
and cannot easily be interpreted. A full Bayesian treatment of a model
is required to produce useful, normalized and calibrated probabilities
for each prediction.

Many interesting models do not have a closed form expression for the
posterior distributions from which predictions are drawn. This
predicament often leads practitioners to turn to powerful MCMC
techniques like NUTS (Hoffman and Gelman 2011). While NUTS creates good
approximations of the posteriors it's limited in terms of parameter and
data scale, making its application to neural networks suboptimal.
Several attempts have been made to create languages which extend modern
deep learning libraries with a probabilistic framework, e.g, Pyro and
Tensorflow Probability. While powerful, we feel that these languages are
verbose and put extra mental burden onto the practitioner.

We introduce Borch, a deep universal probabilistic programming language
built on PyTorch. It provides a true Pythonic look and feel and can
easily enable deep learning practitioners to include uncertainty
estimates into their models. A process we call \texttt{borchify} turns a
standard PyTorch model into its Bayesian counterpart. This process can
operate on the entire network or be constrained to parts of it. This
allows the practitioner to choose which pieces of a neural network that
make the most sense to make Bayesian.

With this work we aim to remove the obstacles deep learning
practitioners have in implementing Bayesian models by providing a
language that looks and feels exactly like PyTorch. We hope that the
community will build upon this and focus more on OOD predictive quality
in all application areas.

\hypertarget{references}{%
\section*{References}\label{references}}
\addcontentsline{toc}{section}{References}

\hypertarget{refs}{}
\begin{CSLReferences}{1}{0}
\leavevmode\vadjust pre{\hypertarget{ref-bakarji2022}{}}%
Bakarji, Joseph, Kathleen Champion, J. Nathan Kutz, and Steven L.
Brunton. 2022. {``Discovering Governing Equations from Partial
Measurements with Deep Delay Autoencoders.''} arXiv.
\url{https://doi.org/10.48550/ARXIV.2201.05136}.

\leavevmode\vadjust pre{\hypertarget{ref-bassett2018maximum}{}}%
Bassett, Robert, and Julio Deride. 2018. {``Maximum a Posteriori
Estimators as a Limit of Bayes Estimators.''} \emph{Mathematical
Programming} 174 (1-2): 129--44.
\url{https://doi.org/10.1007/s10107-018-1241-0}.

\leavevmode\vadjust pre{\hypertarget{ref-bingham2019pyro}{}}%
Bingham, Eli, Jonathan P. Chen, Martin Jankowiak, Fritz Obermeyer,
Neeraj Pradhan, Theofanis Karaletsos, Rohit Singh, Paul Szerlip, Paul
Horsfall, and Noah D. Goodman. 2019. {``Pyro: Deep Universal
Probabilistic Programming.''} \emph{Journal of Machine Learning
Research} 20 (28): 1--6. \url{http://jmlr.org/papers/v20/18-403.html}.

\leavevmode\vadjust pre{\hypertarget{ref-blei2017}{}}%
Blei, David M., Alp Kucukelbir, and Jon D. McAuliffe. 2017.
{``Variational Inference: A Review for Statisticians.''} \emph{Journal
of the American Statistical Association} 112 (518): 859--77.
\url{https://doi.org/10.1080/01621459.2017.1285773}.

\leavevmode\vadjust pre{\hypertarget{ref-bojarski2016end}{}}%
Bojarski, Mariusz, Davide Del Testa, Daniel Dworakowski, Bernhard
Firner, Beat Flepp, Prasoon Goyal, Lawrence D. Jackel, et al. 2016.
{``End to End Learning for Self-Driving Cars.''} arXiv.
\url{https://doi.org/10.48550/ARXIV.1604.07316}.

\leavevmode\vadjust pre{\hypertarget{ref-Brooks2011}{}}%
Brooks, Steve, Andrew Gelman, Galin L. Jones, and Xiao-Li Meng. 2011.
{``{Handbook of Markov Chain Monte Carlo}.''} \emph{Handbook of Markov
Chain Monte Carlo}. \url{https://doi.org/10.1201/b10905}.

\leavevmode\vadjust pre{\hypertarget{ref-debuitleir2013a}{}}%
Buitléir, Amy de, Michael Russell, and Mark Daly. 2013. {``A Functional
Approach to Neural Networks.''} Edited by Edward Z. Yang. \emph{The
Monad.Reader} 21 (March): 5--24.
\url{http://themonadreader.files.wordpress.com/2013/03/issue214.pdf}.

\leavevmode\vadjust pre{\hypertarget{ref-chollet2015keras}{}}%
Chollet, Francois et al. 2015. {``Keras.''} GitHub. 2015.
\url{https://github.com/fchollet/keras}.

\leavevmode\vadjust pre{\hypertarget{ref-dasgupta2020a}{}}%
Dasgupta, Ishita, Eric Schulz, Joshua B Tenenbaum, and Samuel J
Gershman. 2020. {``A Theory of Learning to Infer.''} \emph{Psychological
Review} 127 (3): 412.

\leavevmode\vadjust pre{\hypertarget{ref-davies2021advancing}{}}%
Davies, Alex, Petar Veličković, Lars Buesing, Sam Blackwell, Daniel
Zheng, Nenad Tomašev, Richard Tanburn, et al. 2021. {``Advancing
Mathematics by Guiding Human Intuition with AI.''} \emph{Nature} 600
(7887): 70--74.

\leavevmode\vadjust pre{\hypertarget{ref-davis2021deep}{}}%
Davis, Ernest. 2021. {``Deep Learning and Mathematical Intuition: A
Review of (Davies Et Al. 2021).''} arXiv.
\url{https://doi.org/10.48550/ARXIV.2112.04324}.

\leavevmode\vadjust pre{\hypertarget{ref-deng2012the}{}}%
Deng, Li. 2012. {``The Mnist Database of Handwritten Digit Images for
Machine Learning Research.''} \emph{IEEE Signal Processing Magazine} 29
(6): 141--42.

\leavevmode\vadjust pre{\hypertarget{ref-dillon2017tensorflow}{}}%
Dillon, Joshua V., Ian Langmore, Dustin Tran, Eugene Brevdo, Srinivas
Vasudevan, Dave Moore, Brian Patton, Alex Alemi, Matt Hoffman, and Rif
A. Saurous. 2017. {``TensorFlow Distributions.''} arXiv.
\url{https://doi.org/10.48550/ARXIV.1711.10604}.

\leavevmode\vadjust pre{\hypertarget{ref-farahani2021a}{}}%
Farahani, Abolfazl, Behrouz Pourshojae, Khaled Rasheed, and Hamid R.
Arabnia. 2021. {``A Concise Review of Transfer Learning.''} arXiv.
\url{https://doi.org/10.48550/ARXIV.2104.02144}.

\leavevmode\vadjust pre{\hypertarget{ref-gawlikowski2021a}{}}%
Gawlikowski, Jakob, Cedrique Rovile Njieutcheu Tassi, Mohsin Ali,
Jongseok Lee, Matthias Humt, Jianxiang Feng, Anna Kruspe, et al. 2021.
{``A Survey of Uncertainty in Deep Neural Networks.''} arXiv.
\url{https://doi.org/10.48550/ARXIV.2107.03342}.

\leavevmode\vadjust pre{\hypertarget{ref-Ghahramani2015}{}}%
Ghahramani, Zoubin. 2015. {``Probabilistic Machine Learning and
Artificial Intelligence.''} \emph{Nature} 521 (7553): 452--59.
\url{https://doi.org/10.1038/nature14541}.

\leavevmode\vadjust pre{\hypertarget{ref-goodman2014the}{}}%
Goodman, Noah D, and Andreas Stuhlmüller. 2014. {``{The Design and
Implementation of Probabilistic Programming Languages}.''}
\url{http://dippl.org}.

\leavevmode\vadjust pre{\hypertarget{ref-henriksson2021performance}{}}%
Henriksson, Jens, Christian Berger, Markus Borg, Lars Tornberg, Sankar
Raman Sathyamoorthy, and Cristofer Englund. 2021. {``Performance
Analysis of Out-of-Distribution Detection on Trained Neural Networks.''}
\emph{Information and Software Technology} 130 (February): 106409.
\url{https://doi.org/10.1016/j.infsof.2020.106409}.

\leavevmode\vadjust pre{\hypertarget{ref-hoffman2011the}{}}%
Hoffman, Matthew D., and Andrew Gelman. 2011. {``The No-u-Turn Sampler:
Adaptively Setting Path Lengths in Hamiltonian Monte Carlo.''} arXiv.
\url{https://doi.org/10.48550/ARXIV.1111.4246}.

\leavevmode\vadjust pre{\hypertarget{ref-howard2019searching}{}}%
Howard, Andrew, Mark Sandler, Grace Chu, Liang-Chieh Chen, Bo Chen,
Mingxing Tan, Weijun Wang, et al. 2019. {``Searching for MobileNetV3.''}
arXiv. \url{https://doi.org/10.48550/ARXIV.1905.02244}.

\leavevmode\vadjust pre{\hypertarget{ref-jospin2022hands}{}}%
Jospin, Laurent Valentin, Hamid Laga, Farid Boussaid, Wray Buntine, and
Mohammed Bennamoun. 2022. {``Hands-on Bayesian Neural Networks --- a
Tutorial for Deep Learning Users.''} \emph{IEEE Computational
Intelligence Magazine} 17 (2): 29--48.
\url{https://doi.org/10.1109/MCI.2022.3155327}.

\leavevmode\vadjust pre{\hypertarget{ref-kiureghian2009aleatory}{}}%
Kiureghian, Armen Der, and Ove Ditlevsen. 2009. {``Aleatory or
Epistemic? Does It Matter?''} \emph{Structural Safety} 31 (2): 105--12.
https://doi.org/\url{https://doi.org/10.1016/j.strusafe.2008.06.020}.

\leavevmode\vadjust pre{\hypertarget{ref-klas2018uncertainty}{}}%
Kläs, Michael, and Anna Maria Vollmer. 2018. {``Uncertainty in Machine
Learning Applications: A Practice-Driven Classification of
Uncertainty.''} In \emph{Computer Safety, Reliability, and Security},
edited by Barbara Gallina, Amund Skavhaug, Erwin Schoitsch, and
Friedemann Bitsch, 431--38. Cham: Springer International Publishing.

\leavevmode\vadjust pre{\hypertarget{ref-li2022competition}{}}%
Li, Yujia, David Choi, Junyoung Chung, Nate Kushman, Julian
Schrittwieser, Rémi Leblond, Tom Eccles, et al. 2022.
{``Competition-Level Code Generation with AlphaCode.''} arXiv.
\url{https://doi.org/10.48550/ARXIV.2203.07814}.

\leavevmode\vadjust pre{\hypertarget{ref-paisley2012variational}{}}%
Paisley, John, David Blei, and Michael Jordan. 2012. {``Variational
Bayesian Inference with Stochastic Search.''} arXiv.
\url{https://doi.org/10.48550/ARXIV.1206.6430}.

\leavevmode\vadjust pre{\hypertarget{ref-NEURIPS2019_9015}{}}%
Paszke, Adam, Sam Gross, Francisco Massa, Adam Lerer, James Bradbury,
Gregory Chanan, Trevor Killeen, et al. 2019. {``PyTorch: An Imperative
Style, High-Performance Deep Learning Library.''} In \emph{Advances in
Neural Information Processing Systems 32}, edited by H. Wallach, H.
Larochelle, A. Beygelzimer, F. dAlché-Buc, E. Fox, and R. Garnett,
8024--35. Curran Associates, Inc.
\url{http://papers.neurips.cc/paper/9015-pytorch-an-imperative-style-high-performance-deep-learning-library.pdf}.

\leavevmode\vadjust pre{\hypertarget{ref-salvatier2016probabilistic}{}}%
Salvatier, John, Thomas V Wiecki, and Christopher Fonnesbeck. 2016.
{``Probabilistic Programming in Python Using PyMC3.''} \emph{PeerJ
Computer Science} 2: e55.

\leavevmode\vadjust pre{\hypertarget{ref-shwartz-ziv2021tabular}{}}%
Shwartz-Ziv, Ravid, and Amitai Armon. 2021. {``Tabular Data: Deep
Learning Is Not All You Need.''} arXiv.
\url{https://doi.org/10.48550/ARXIV.2106.03253}.

\leavevmode\vadjust pre{\hypertarget{ref-siddharth2017learning}{}}%
Siddharth, N., Brooks Paige, Jan-Willem van de Meent, Alban Desmaison,
Noah Goodman, Pushmeet Kohli, Frank Wood, and Philip H.S. Torr. 2017.
{``Learning Disentangled Representations with Semi-Supervised Deep
Generative Models.''} In \emph{Advances in Neural Information Processing
Systems}, edited by I. Guyon, U. V. Luxburg, S. Bengio, H. Wallach, R.
Fergus, S. Vishwanathan, and R. Garnett, 5925--35. Neural Information
Processing Systems.

\leavevmode\vadjust pre{\hypertarget{ref-SDT2018}{}}%
Stan Development Team. 2018. {``{RStan}: The {R} Interface to {Stan}.''}
\url{http://mc-stan.org/}.

\leavevmode\vadjust pre{\hypertarget{ref-tolpin2016design}{}}%
Tolpin, David, Jan-Willem van de Meent, Hongseok Yang, and Frank Wood.
2016. {``Design and Implementation of Probabilistic Programming Language
Anglican.''} In \emph{Proceedings of the 28th Symposium on the
Implementation and Application of Functional Programming Languages}. IFL
2016. New York, NY, USA: Association for Computing Machinery.
\url{https://doi.org/10.1145/3064899.3064910}.

\leavevmode\vadjust pre{\hypertarget{ref-tran2016edward}{}}%
Tran, Dustin, Alp Kucukelbir, Adji B. Dieng, Maja Rudolph, Dawen Liang,
and David M. Blei. 2016. {``Edward: A Library for Probabilistic
Modeling, Inference, and Criticism.''} arXiv.
\url{https://doi.org/10.48550/ARXIV.1610.09787}.

\leavevmode\vadjust pre{\hypertarget{ref-zheng2016improving}{}}%
Zheng, Stephan, Yang Song, Thomas Leung, and Ian Goodfellow. 2016.
{``Improving the Robustness of Deep Neural Networks via Stability
Training.''} arXiv. \url{https://doi.org/10.48550/ARXIV.1604.04326}.

\end{CSLReferences}

\end{document}